\documentclass[conference]{IEEEtran}
\IEEEoverridecommandlockouts
\usepackage{cite}
\usepackage{amsmath,amssymb,amsfonts}
\usepackage{algorithm}
\usepackage{algpseudocode}
\usepackage[dvipdfmx]{graphicx}
\usepackage{subcaption}
\usepackage{textcomp}
\usepackage{xcolor}
\def\BibTeX{{\rm B\kern-.05em{\sc i\kern-.025em b}\kern-.08em
    T\kern-.1667em\lower.7ex\hbox{E}\kern-.125emX}}
    
\usepackage[top=0.7in,bottom=1.1in,left=0.7in,right=0.7in]{geometry}

\usepackage{mathtools}

\usepackage{optidef}
\usepackage[super]{nth}

\DeclareMathOperator*{\argmax}{arg\,max}
\DeclareMathOperator*{\argmin}{arg\,min}
\newcommand{\E}{\mathbb{E}}

\usepackage{flushend}
\usepackage{stfloats}

\newcommand{\markupdraft}[2]{
    \ifthenelse{\equal{#1}{display}}{#2}{}
    \ifthenelse{\equal{#1}{color}}{\color{#2}}{}
}

\newcommand{\newcolored}[3][]{{\markupdraft{color}{#2}#3}
    \ifthenelse{\equal{#1}{}}{}{\markupdraft{display}{{\color{yellow!70!black}[#1]}}}} 

\begin{document}

\title{Neural Architecture Search for Improving Latency-Accuracy Trade-off in Split Computing\\
\thanks{This work was supported in part by JST PRESTO under Grant JPMJPR2035 and JPMJPR2133, and a project commissioned by NEDO (JPNP18002).}
}

\author{\IEEEauthorblockN{
Shoma Shimizu\IEEEauthorrefmark{1}, 
Takayuki Nishio\IEEEauthorrefmark{2},
Shota Saito\IEEEauthorrefmark{1}\,\IEEEauthorrefmark{3}, 
Yoichi Hirose\IEEEauthorrefmark{1},
Chen Yen-Hsiu\IEEEauthorrefmark{1}, 
Shinichi Shirakawa\IEEEauthorrefmark{1}
}
\IEEEauthorblockA{\IEEEauthorrefmark{1}\textit{Graduate School of Environment and Informations Sciences} \\
\textit{Yokohama National University}, Yokohama, Japan \\
\{shimizu-shoma-kr, saito-shota-bt, hirose-youichi-kc, chen-yen-hsiu-tx\}@ynu.jp, shirakawa-shinichi-bg@ynu.ac.jp}
\IEEEauthorblockA{\IEEEauthorrefmark{2}\textit{School of Engineering, Tokyo Institute of Technology}, Tokyo, Japan \\
nishio@ict.e.titech.ac.jp}
\IEEEauthorblockA{\IEEEauthorrefmark{3}\textit{SkillUp AI, Co., Ltd.}, Tokyo, Japan}
}

\maketitle

\begin{abstract}
This paper proposes a neural architecture search (NAS) method for split computing. Split computing is an emerging machine-learning inference technique that addresses the privacy and latency challenges of deploying deep learning in IoT systems. In split computing, neural network models are separated and cooperatively processed using edge servers and IoT devices via networks. Thus, the architecture of the neural network model significantly impacts the communication payload size, model accuracy, and computational load.
In this paper, we address the challenge of optimizing neural network architecture for split computing. To this end, we proposed NASC, which jointly explores optimal model architecture and a split point to achieve higher accuracy while meeting latency requirements (i.e., smaller total latency of computation and communication than a certain threshold). NASC employs a one-shot NAS that does not require repeating model training for a computationally efficient architecture search.
Our performance evaluation using hardware (HW)-NAS-Bench of benchmark data demonstrates that the proposed NASC can improve the ``communication latency and model accuracy" trade-off, i.e., reduce the latency by approximately 40--60\% from the baseline, with slight accuracy degradation.


\end{abstract}


\section{Introduction}\label{sec:intro}
Combining the physical sensing of IoT devices with deep learning-based data analysis has been gained considerable attention to enable multiple novel applications. However, the strict privacy and latency demands of IoT applications pose challenges. For example, IoT sensors (e.g., visual and audio sensors) in smart home applications obtain privacy-sensitive data that should not be exposed \cite{Lin16}, and the latency requirements for the factory automation and smart grids are less than 10 ms and 20 ms, respectively \cite{Schulz17}. 

Split computing, which provides an intermediate option between edge computing and local computing, has been proposed to address the privacy and latency challenges of deploying deep learning in IoT systems. 
In the split computing, a well-trained deep neural network (DNN) is divided into sub-DNNs, and the IoT devices and the edge server collaboratively execute the sub-DNNs by exchanging information (e.g., intermediate outputs) via IoT networks.
Because the computation to execute the DNN is partially offloaded to the edge server, and the privacy-sensitive data are processed and converted into an intermediate representation, the split computing can reduce the computation latency and data leakage risk in the DNN-based IoT applications.

However, split computing poses a new research challenge: the ``communication latency vs. model accuracy" trade-off.
The payload size of the intermediate output is typically larger than that of the raw input or output of the original model, which requires introducing a ``bottleneck structure" between the head and tail networks. Narrowing the bottleneck allows for a smaller payload size but may result in significant performance degradation during inference. This is the ``communication latency vs. model accuracy" trade-off.

Many studies have been conducted to improve the ``communication latency vs. model accuracy" trade-off. Eshratifar et al. have studied the bottleneck injection, called BottleNet, and demonstrated that BottleNet can improve end-to-end latency and reduce mobile energy consumption compared with the cloud-based computation without significant degradation of model accuracy \cite{Eshratifar19}.
Matsubara et al. have proposed a method to train the head network to maintain the model accuracy, called head network distillation \cite{Matsubara20}. Head network distillation employs knowledge distillation to transfer the knowledge of the head network generated from a well-trained original DNN into a compressed head network employing a bottleneck structure. 
Itahara et al. have proposed a COMtune, which can improve the model's robustness against lossy compression of intermediate outputs and packet loss in the IoT networks \cite{Itahara22}. 

As also mentioned in the survey \cite{Survey_matsubara}, although many studies have investigated split computing, many research challenges still need to be addressed. In this study, we focused on optimizing head and tail network design. The previous works used a handmade architecture of the head and tail networks. However, the architecture of the head and tail networks is of considerable importance in split computing because they have a significant impact on the payload size, model accuracy, and computation load. The smaller the bottleneck structure, the less traffic there is, but the lower the model accuracy. Moreover, because IoT devices generally have limited computational resources, the head network must be lightweight enough to satisfy latency constraints. Although few studies have addressed the optimization of the split point and bottleneck placement \cite{Li18}, further performance improvements can be still achieved by optimizing the overall model structure.

To this end, we propose a neural architecture search (NAS)~\cite{Elsken2019} method for split computing, referred to as NASC. The proposed NASC algorithm is based on one-shot NAS and employs an adaptive stochastic natural gradient (ASNG)~\cite{akimoto19} method as the search algorithm to efficiently search for the optimal model architecture and split point jointly that achieves higher accuracy while meeting latency requirements (i.e., smaller total latency of computation and communication than a certain threshold). One-shot NAS including our method trains the weights of a supernetwork that contains all candidate architectures only once during the search process and can drastically reduce the search cost. 
We conducted a performance evaluation using hardware (HW)-NAS-Bench \cite{HW-NAS}, which is benchmark data obtained using resource-constrained devices, and demonstrated that the proposed NASC can reduce the latency to less than a certain threshold with slight accuracy degradation.

\section{NASC: NAS for Split Computing}
We propose a method for obtaining a model architecture that achieves high accuracy with low latency in the split computing via lossy wireless links. In the following sections, we first summarize several assumptions regarding our proposal and then present NASC in more detail.

\subsection{Assumptions}
We consider a networked computing system consisting of a cloud server, an edge server, and end devices. The cloud server searches for and trains a deep neural network that works well in split computing. The edge server and devices collaboratively conduct inferences with the sensing data obtained by the devices and the model trained by the cloud server in a split computing manner.

A trained neural network model is split into a head and tail networks in split computing. The output of the head network (i.e., the intermediate representation) is transmitted from the device to the edge server via a wireless network. In this study, we assume that the wireless link is stable but unreliable; that is, the throughput for transmitting the model output does not change, but the packet can be dropped problematically, which can cause part of the intermediate representation to be missing.

Moreover, we assume that the computational power of end devices and throughput between the edge server and devices are limited because of resource-constrained IoT devices and networks. This causes non-negligible latency in calculations at the end devices and data transfer between the edge server and devices.

\subsection{Mathematical Formulation} \label{sec:formulation}
As with existing studies on NAS, we address the following optimization problem:
\begin{mini}
{\substack{x\in\mathcal{X}, a\in\mathcal{A}}}
{f(x, a) \enspace,}{}{}
\end{mini}
where $f: \mathcal{X} \times \mathcal{A} \rightarrow \mathbb{R}$ is an objective function, such as a loss function, $x \in \mathcal{X}$ and $a \in \mathcal{A}$ are the weight and architecture parameters of a neural network model, respectively.

Let the neural network model be defined as 
$N(\cdot|x, a) = N_\mathrm{h}(\cdot | x_\mathrm{h}, a_\mathrm{h}) \circ N_\mathrm{t}(\cdot | x_\mathrm{t}, a_\mathrm{t})$, where $N_\mathrm{h}(\cdot | x_\mathrm{h}, a_\mathrm{h})$ and $N_\mathrm{t}(\cdot | x_\mathrm{t}, a_\mathrm{t})$ denote the head and tail network parameterized by $(x_\mathrm{h}, a_\mathrm{h})$ and $(x_\mathrm{t}, a_\mathrm{t})$, respectively. $f(\cdot) \circ g(\cdot)$ denotes the composite function of $f (\cdot)$ and $g(\cdot)$.
Thus, the optimization problem NASC addresses can be written as
\begin{mini}
{\substack{x_\mathrm{h}, x_\mathrm{t} \in\mathcal{X}, a_\mathrm{h}, a_\mathrm{t} \in \mathcal{A}}}
{f(x_\mathrm{h}, a_\mathrm{h}, x_\mathrm{t}, a_\mathrm{t})}{}{} \enspace.
\end{mini}

In contrast to the existing NAS problem where model loss is minimized, NASC must consider computation and communication latency due to resource-constrained devices. We define the computational latency of a model $N(\cdot|x,a)$ processed by computing node $i$ as $T^\mathrm{comp}_i(a)$. Because the computational processes of the head and tail networks are independent, the computation latency for the head and tail networks can be written as $T^\mathrm{comp}_i(a_\mathrm{h})$ and $T^\mathrm{comp}_j(a_\mathrm{t})$, respectively. In split computing, the node processing tail network is often an edge or cloud server with much higher computation power than end devices such as Raspberry Pi and Jetson. Thus, searching $a_\mathrm{h}$ that can achieve low latency without significantly degrading the accuracy is essential.

On one hand, we define the communication latency of split computing with $N_\mathrm{h}(\cdot | x_\mathrm{h}, a_\mathrm{h})$ and $N_\mathrm{t}(\cdot | x_\mathrm{t}, a_\mathrm{t})$ conducted by computing nodes $i$ and $j$ as $T^\mathrm{comm}_{i,j}(x_\mathrm{h}, a_\mathrm{h}, x_\mathrm{t}, a_\mathrm{t})$.  Communication latency mainly depends on the communication throughput between computing node $i$ and $j$ and the data size of the output of the head network $N_\mathrm{h}(\cdot | x_\mathrm{h}, a_\mathrm{h})$. Because a larger data size increases the communication latency, we need to find $a_\mathrm{h}$ to achieve small output size (i.e., bottleneck architecture) without degrading accuracy. Moreover, as reported in \cite{Itahara22}, in the same model architecture, model training that incorporates dropout can suppress accuracy degradation when data are dropped on the communication channel in split computing, which enables the communication system to employ less reliable but fewer latency protocols (i.e., UDP).

The end-to-end latency of the split computing with nodes $i$ and $j$ is then written as
\begin{align}
T = T^\mathrm{comp}_i(a_\mathrm{h}) + T^\mathrm{comm}_{i,j}(x_\mathrm{h}, a_\mathrm{h}, x_\mathrm{t}, a_\mathrm{t}) + T^\mathrm{comp}_j(a_\mathrm{t}) \enspace.
\end{align}
As mentioned in Sect.~\ref{sec:intro}, the objective of NASC is to improve the trade-off between the latency and model accuracy. To reduce communication latency, we must introduce a narrow bottleneck architecture that may induce accuracy degradation. The accuracy degradation caused by the bottleneck may be compensated by the large head and tail networks, but it will increase the computation latency. Moreover, the effect of communications on model accuracy, that is, accuracy degradation due to packet loss, must also be considered. 

In this paper, we introduce a penalty that increases when the threshold is exceeded and define the objective function as a weighted sum of the model loss function and latency penalty. 
Finally, the optimization problem solved in NASC is written as follows:
\begin{mini!}|l|
{\substack{x_\mathrm{h}, x_\mathrm{t} \in\mathcal{X},\\ a_\mathrm{h}, a_\mathrm{t} \in\mathcal{A}}}
{\epsilon_\mathrm{loss}l_\mathrm{SC}(x_\mathrm{h}, a_\mathrm{h}, x_\mathrm{t}, a_\mathrm{t})+\epsilon_\mathrm{lat}\tau}{}{} \label{eq:opt}
\addConstraint{\tau}{= \max(0, T-T_\mathrm{th})}
\addConstraint{T}{= T^\mathrm{comp}_i(a_\mathrm{h}) + T^\mathrm{comp}_j(a_\mathrm{t}) + T^\mathrm{comm}_{i,j}}
\end{mini!}
where $\epsilon_\mathrm{loss}$ and $\epsilon_\mathrm{lat}$ denote the weights for model loss and latency penalty, respectively, $l_\mathrm{SC}$ denotes the loss function of the split computing model when causing packet loss, $T_\mathrm{th}$ is the threshold for the latency causing the penalty, and $T^\mathrm{comm}_{i,j}$ is a shorthand notation for $T^\mathrm{comm}_{i,j}(x_\mathrm{h}, a_\mathrm{h}, x_\mathrm{t}, a_\mathrm{t})$. 

\subsection{Algorithm}
To optimize the weight and architecture parameters in the loss function \eqref{eq:opt}, we applied the one-shot NAS method based on an adaptive stochastic natural gradient neural architecture search (ASNG-NAS)\cite{akimoto19}. ASNG-NAS considers the probability distribution $P_{\theta}(a)$ that generates the architecture parameters, and we optimize the weight parameters $x$ and the distribution parameters $\theta$ by minimizing the expected objective function $\E_{P_{\theta}(a)}\left\lbrack f(x, a)\right\rbrack$. More details on the ASNG-NAS can be found in \cite{akimoto19}.

For solving NASC using the ASNG-NAS, we consider optimizing the weight and architecture parameters of the large neural network before splitting it, rather than optimizing the head and tail networks separately. Therefore, we introduce a split point $k \in \mathcal{K}$ and redefine the neural network model as $N(\cdot|x, a, k)$. The split point $k$ represents the layer or block number, and the model $N$ is divided into the head and tail networks based on $k$. We aim to optimize $x, a$ and $k$ based on the ASNG-NAS. The split point $k$ is sampled from the probability distribution $P_{\theta}$ as well as the architecture parameters. Therefore, we denote $\boldsymbol{a} = (a, k) \in \mathcal{A} \times \mathcal{K}$ as a random variable that follows $P_{\theta}(\boldsymbol{a})$, and the expected objective function of ASNG-NAS for NASC is written as $J(x, \theta) = \E_{P_{\theta}(\boldsymbol{a})} \lbrack \epsilon_\mathrm{loss}l_\mathrm{SC}(x, \boldsymbol{a})+\epsilon_\mathrm{lat}\tau \rbrack$. In this paper, we use $l_{\mathrm{SC}}(x, \boldsymbol{a}) = \sum_{p \in \mathcal{P}}\mathcal{L}(x, \boldsymbol{a}, p)/|\mathcal{P}|$ where $\mathcal{L}(x, \boldsymbol{a}, p)$ is a loss function such as cross-entropy by inserting a dropout at the split point $k$ using a dropout rate $p$ and $\mathcal{P} = \{0.0, 0.1, 0.2, 0.3, 0.4, 0.5\}$ is a set of dropout rates, and $\epsilon_\mathrm{loss} = \epsilon_\mathrm{lat} = 1$.

\begin{algorithm}[t]
\caption{ASNG-NAS for NASC}\label{alg:nasc}
\begin{algorithmic}[1]
\Require $\alpha=1.5, \delta_{\theta}^{0}=1, \lambda_{x} = \lambda_{\theta} = 2$
\State initialize the weight parameters $x$ and the distribution parameters $\theta$
\Repeat
\Comment{weight pre-training}
\State sample $\lambda_{x}$ pair of architecture and split point from a uniform distribution and update $x$ using \eqref{grad_x}
\Until{termination conditions are met}
\State $\Delta=1, \gamma=0, \boldsymbol{s}=\mathbf{0}, t=0$
\Repeat
\Comment{distribution update}
\State $\delta_{\theta}=\delta_{\theta}^{0} / \Delta, \beta=\delta_{\theta} / n_{\theta}^{1 / 2}$
\State update $\theta^{t+1}$ with \eqref{upd_th}, then force $\theta^{t+1} \in \Theta$ by projection
\State $\epsilon_{\theta}=\delta_{\theta} /\left\|G\left(\theta^{t}\right)\right\|_{\mathbf{F}\left(\theta^{t}\right)}$ \label{adaptive_lr}
\State $\boldsymbol{s} \leftarrow(1-\beta) \boldsymbol{s}+\sqrt{\beta(2-\beta)} 
\frac{
\mathbf{F}\left(\theta^{t}\right)^{\frac{1}{2}} G\left(\theta^{t}\right)
}{\left\|G\left( \theta^{t}\right)\right\|_{\mathbf{F}\left(\theta^{t}\right)}}
$
\State $\gamma \leftarrow(1-\beta)^{2} \gamma+\beta(2-\beta)$
\State $\Delta \leftarrow \min \left(\Delta_{\max}, \Delta \exp \left(\beta\left(\gamma-\|\boldsymbol{s}\|^{2} / \alpha\right)\right)\right)$
\Until{termination conditions are met}
\State  sample the most likely architecture and split point $\boldsymbol{a}^{*} = \argmax_{\boldsymbol{a}} P_{\theta}(\boldsymbol{a})$, and update the weight parameters
\Comment{weight re-training}
\end{algorithmic}
\end{algorithm}

Algorithm \ref{alg:nasc} summarizes the proposed method.
To run the algorithm, we set a supernet, a model whose all sub-models are equivalent to all the models in $\mathcal{A}$. The architecture and split point are represented by the following $D$ dimensional categorical variable: $\boldsymbol{\alpha} = (\alpha_1, \alpha_2, \dots, \alpha_D)$ where $\alpha_d$ possesses $K_d$ categories. Note that $\alpha_1, \dots, \alpha_{D-1}$ and $\alpha_D$ are tied to the architecture and split point, respectively. We treat architecture and split point as one-hot vectors $\boldsymbol{a} = (a_1, a_2, \dots a_D)$ where $a_d = (a_{d, 1}, a_{d, 2}, \dots a_{d, K_d})^{\mathrm{T}} \in \{0, 1\}^{K_d}$ such that $\sum^{K_d}_{k=1}a_{d, k} = 1$.
Also, we set a $D$ dimensional categorical distribution $P_{\theta}(\boldsymbol{a}) = \prod^{D}_{d=1}\prod^{K_d}_{k=1}(\theta_{d, k})^{a_{d, k}}$ where $\theta_{d, k} \in [0, 1]$ is the probability of being $a_{d, k} = 1$ such that $\sum^{K_d}_{k=1} \theta_{d, k} = 1$, and the distribution parameters is represented by $\theta = (\theta_1, \theta_2, \dots, \theta_D) \in \Theta$ where $\theta_d = (\theta_{d, 1}, \theta_{d, 2}, \dots, \theta_{d, K_d})^{\mathrm{T}}$, and the number of distribution parameters is $n_\theta = \sum^{D}_{d=1}\sum^{K_d}_{k=1}1$.\footnote{We can omit the distribution parameter for the last categorical element owing to the condition of $\sum^{K_d}_{k=1} \theta_{d, k} = 1$.}

The proposed algorithm consists of three stages; weight pre-training, distribution update\footnote{The original ASNG-NAS alternates between updating weight and distribution parameters. However, we observed improved results in preliminary experiments when the weights and distribution parameters were updated sequentially, as in \cite{Noda2022}.}, and weight re-training. 

\vspace{3mm}
\noindent \textbf{Weight Pre-training:}\quad
In the weight pre-training stage, we optimize the weight parameters $x$ in the supernet to minimize $J(x, \theta)$. The gradient of weight parameters is given by $\nabla_x J(x, \theta) = \mathbb{E}_{P_{\theta}(\boldsymbol{a})} \lbrack \nabla_x \epsilon_{\mathrm{loss}} l_{\mathrm{SC}}(x, \boldsymbol{a}) \rbrack$. In most cases, it is difficult to compute analytically $\nabla_x J(x, \theta)$. In addition, computing $l_{\mathrm{SC}}$ requires $|\mathcal{P}|$ times forward propagation, which is expensive for computation cost. Therefore, $\nabla_x J(x, \theta)$ is approximated using Monte-Carlo method with $\lambda_{x} (=2)$ samples $\boldsymbol{a}^{(1)}, \boldsymbol{a}^{(2)}, \dots, \boldsymbol{a}^{(\lambda_x)}$ sampled from the uniform distribution and $l_{\mathrm{SC}}(x, \boldsymbol{a}) \approx \mathcal{L}(x, \boldsymbol{a}, p=0.5)$ to reduce the computation cost of training the weights (the same applies to the re-training stage). The approximated gradient of weight parameters is given by
\begin{align}
G(x^{t+1}) &= \frac{1}{\lambda_{x}} \sum_{i=1}^{\lambda_{x}}\epsilon_{\mathrm{loss}} \nabla_{x} \mathcal{L}(x^{t}, \boldsymbol{a}^{(i)}, p=0.5) \label{grad_x} \enspace,
\end{align}
where $t$ is a time step. We note that the weight parameters $x$ can be updated using any stochastic gradient descent (SGD) method with \eqref{grad_x}.
We repeat this process for 30 epochs.

\vspace{3mm}
\noindent \textbf{Distribution Update:}\quad
In the distribution update stage, we in turn optimize the distribution parameters $\theta$ under the trained weight parameters $x^{*}$. We use the natural gradient $\tilde{\nabla}_\theta J(x^{*}, \theta) = \mathbb{E}_{P_{\theta}(\boldsymbol{a})} \lbrack (\epsilon_{\mathrm{loss}} l_{\mathrm{SC}}(x^{*}, \boldsymbol{a}) +\epsilon_\mathrm{lat} \tau) \tilde{\nabla}_\theta \log P_{\theta}(\boldsymbol{a}) \rbrack$ to update the distribution parameters. Here, $\tilde{\nabla}_\theta = \boldsymbol{\mathrm{F}}(\theta)^{-1}\nabla_\theta$ where $\boldsymbol{\mathrm{F}}(\theta)$ is the Fisher information matrix (FIM), and $\tilde{\nabla}_\theta\log P_{\theta}(\boldsymbol{a})$ is given by $a - \theta$ under the categorical distribution \cite{akimoto19}. However, $\tilde{\nabla}_\theta J(x^{*}, \theta)$ is difficult to compute analytically as with the weight parameters. Therefore, $\tilde{\nabla}_\theta J(x^{*}, \theta)$ is also approximated using Monte-Carlo method with $\lambda_{\theta} (=2)$ samples, and the distribution parameters is updated as following:
\begin{align} 
G(\theta^{t}) &= \frac{1}{\lambda_{\theta}}\sum_{i=1}^{\lambda_{\theta}} u^{(i)}(a^{(i)} - \theta^{t})  \enspace, \\
\theta^{t+1} &= \theta^{t} + \epsilon_{\theta} G(\theta^{t}) \enspace, \label{upd_th}
\end{align}
where $u^{(i)}$ is the utility value based on the objective value $i$-th sample, and $\epsilon_{\theta}$ is an adaptive learning rate updated according to line \ref{adaptive_lr} in Algorithm \ref{alg:nasc}. When $\lambda_{\theta}=2$, the value of the better sample is assigned $2$, and that of the worse sample is $-2$.
As for the termination condition, this study simply set the end of the update at 90 epochs.

\vspace{3mm}
\noindent \textbf{Weight Re-training:}\quad
In the weight re-training stage, we sample the most likely architecture and split point $\boldsymbol{a}^{*} = \argmax_{\boldsymbol{a}} P_{\theta}(\boldsymbol{a})$, and train the weight parameters from scratch to minimize $\mathcal{L}(x, \boldsymbol{a}^{*}, p=0.5)$ for $300$ epochs.

\begin{table}[tb]
    \centering
    \caption{The candidate blocks in the FBNet search space.}
    \begin{tabular}{c|c|c|c}
        \hline
        Block type & expansion ($e$) & Kernel ($K$)  & Group\\ \hline
        k3\_e1 & 1 & 3  & 1\\
        k3\_e1\_g2 & 1 & 3  & 2\\
        k3\_e3 & 3 & 3  & 1\\
        k3\_e6 & 6 & 3  & 1\\
        k5\_e1 & 1 & 5  & 1\\
        k5\_e1\_g2 & 1 & 5  & 2\\
        k5\_e3 & 3 & 5  & 1\\
        k5\_e6 & 6 & 5  & 1\\
        skip & - & -  & -\\
        \hline
    \end{tabular}
    \label{table:candidate_blocks_fbnet}
\end{table}
\begin{table}[tb]
    \centering
    \caption{The additional candidate blocks in the extended search space.}
    \begin{tabular}{c|c|c|c}
        \hline
        Block type & expansion ($e$) & Kernel ($K$) & Group\\ \hline
        k3\_e1/2 & 1/2 & 3  & 1\\
        k3\_e1/4 & 1/4 & 3  & 1\\
        k3\_e1/8 & 1/8 & 3  & 1\\
        k5\_e1/2 & 1/2 & 5  & 1\\
        k5\_e1/4 & 1/4 & 5  & 1\\
        k5\_e1/8 & 1/8 & 5  & 1\\
        \hline
    \end{tabular}
    \label{table:candidate_blocks_original}
\end{table}
\section{Experimental Evaluation}
\subsection{Setups}
\noindent \textbf{Dataset and search space:}\quad
We conducted experiments using the CIFAR-100 dataset and HW-NAS-Bench \cite{HW-NAS}, a hardware performance dataset for hardware-aware NAS. HW-NAS-Bench provides the computation latency and energy cost of all the network architectures in the search spaces of both NAS-Bench-201 \cite{dong2020nasbench201} and FBNet \cite{FBNet} on specific hardware such as Raspberry Pi 4, FPGA, and Edge GPU. In this experiment, we considered the FBNet search space \cite{FBNet} and leveraged the latency performance of Rasberry Pi 4 and Edge GPU as those of an end device and edge server, respectively. 

FBNet search space is a layer-wise search space constructed with a fixed macro architecture that defines the number of layers and each layer's input/output dimensions.
For better accuracy and efficiency, each layer of the network can independently choose different blocks from the candidate block, except for the first and last three layers with fixed operators.
Each candidate block contains three operations: a 1$\times$1 convolution, a K$\times$K depthwise convolution, where K denotes the kernel size, and another 1$\times$1 convolution.
It is important to note that only the first 1$\times$1 convolution and the depthwise convolution are followed by ReLU activation functions, and there is no activation function following the last 1$\times$1 convolution.
Furthermore, to control the block, we use an expansion ratio to determine the input/output channel size expansion of the K$\times$K depthwise convolution.
Each candidate block in the search space can choose a different expansion ratio, kernel size, and the number of groups for the group convolution.  
In the experiment, there were twenty-two layers that we needed to search for from nine predefined candidate blocks, as listed in Table \ref{table:candidate_blocks_fbnet}. The block ``skip'' means that there is no operation in that layer.
Consequently, it contains ${9^{22}} \approx {10^{21}}$ possible architectures in FBNet search space.

\begin{figure}[tb]
    \centering
            \begin{minipage}[t]{0.49\linewidth}
                \centering
                \includegraphics[scale=0.4]{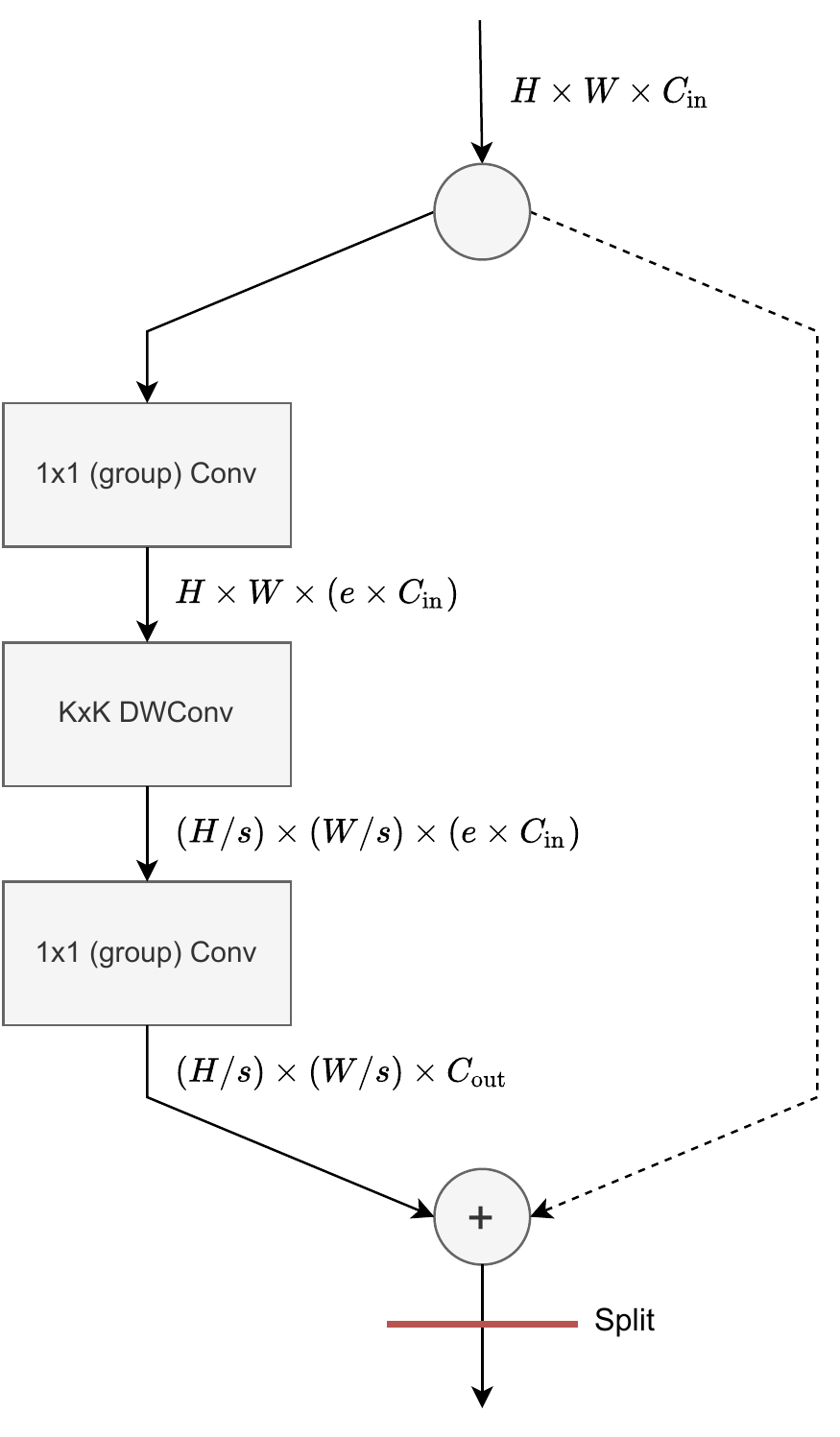}
                \subcaption{The candidate block in the FBNet search space.}
                \label{fig:block_fbnet}
            \end{minipage}
            \begin{minipage}[t]{0.49\linewidth}
                \centering
                \includegraphics[scale=0.5]{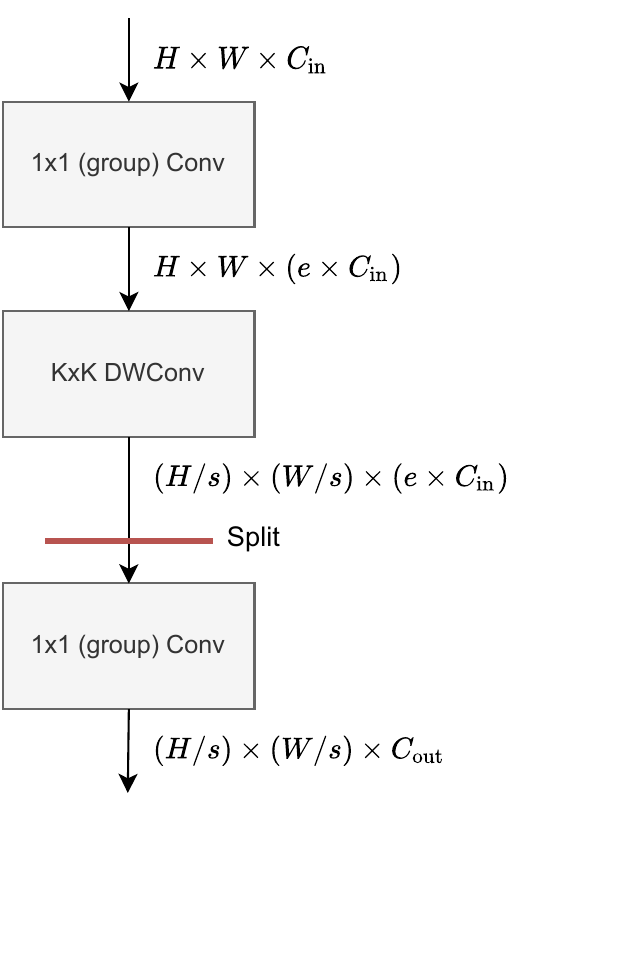}
                \subcaption{The additional candidate block in the extended search space.}
                \label{fig:block_original}
            \end{minipage}
        \caption{The candidate blocks in the FBNet search space and the extended search space. Red lines indicate the split points.}\label{fig:blocks}
\end{figure}
This study also searched for a split point in addition to searching for candidate blocks.
We set 23 candidate split points: 22 after each block and one after the first convolution layer.
Fig. \ref{fig:block_fbnet} shows the candidate block's operations and the split point. The skip connection is inserted when the input and output sizes of the block are the same.
We set $T_{\mathrm{th}} = 30$ (ms) in the FBNet search space.
Note that the FBNet search space originally supported only the ImageNet dataset; however, the work of \cite{HW-NAS} created a macro-architecture for CIFAR-100.

Additionally, we consider an extension of FBNet search space, where the computation latency is approximated from the amount of computation (FLOPs) using HW-NAS-Bench.
The extented search space introduces candidate blocks with an expansion ratio of less than one into the FBNet search space in order to utilize the bottleneck structure.
Table \ref{table:candidate_blocks_original} lists the additional candidate blocks in the extended search space.
The additional blocks change the split point immediately after the K$\times$K depthwise convolution and remove the skip connection to reduce the transfer data size.
Fig. \ref{fig:block_original} shows the additional block's operations and split positions.
We set $T_{\mathrm{th}} = 15$ (ms) in the extended search space.

Because the measured latency cannot be obtained from HW-NAS-Bench owing to the additional blocks, we used the latency estimated using FLOPs instead.
We assume that the device has a specific computation power and that the latency is determined by the model's FLOPs and the device's computation power.
In particular, we assume that the latency $T^{\mathrm{comp}}$ is determined as follows:
\begin{align}
    T^{\mathrm{comp}} = \frac{C^{\mathrm{comp}}}{\Pi}.
\end{align}
Here, $C^{\mathrm{comp}}$ is the model's computation cost (i.e., FLOPs), and $\Pi$ is the device's computation power.
We estimated the computation power using the latency in HW-NAS-Bench using the following equation:
\begin{align}
    \Pi = \argmin_{\pi} \sum_{l=1}^{22} \sum_{k=1}^{9} \left( T_{l,k}^{\mathrm{comp}} - \frac{C_{l,k}^{\mathrm{comp}}}{\pi} \right)^2,
\end{align}
where $T^{\mathrm{comp}}_{l,k}$ and $C^{\mathrm{comp}}_{l,k}$ denote the measured latency and FLOPs of the $k$-th block in the $l$-th layer, respectively.
Table \ref{table:computation_power} lists the devices' estimated computation power.
\begin{table}[tb]
    \centering
    \caption{Devices' estimated computation power}
    \begin{tabular}{c|c}
        \hline
        device & computation power (GFLOPS) \\ \hline
        EdgeGPU & 8.0213 \\
        Raspi4 & 2.3562 \\
        \hline
    \end{tabular}
    \label{table:computation_power}
\end{table}

Note that HW-NAS-Bench~\cite{HW-NAS} points out that the correlation between FLOPs and EdgeGPU latency is small. However, this experiment was conducted to verify that this method can be applied when the latency is obtained in some way and the precision of the latency is not critical.

\begin{figure}
        \centering
        \includegraphics[width=0.7\linewidth]{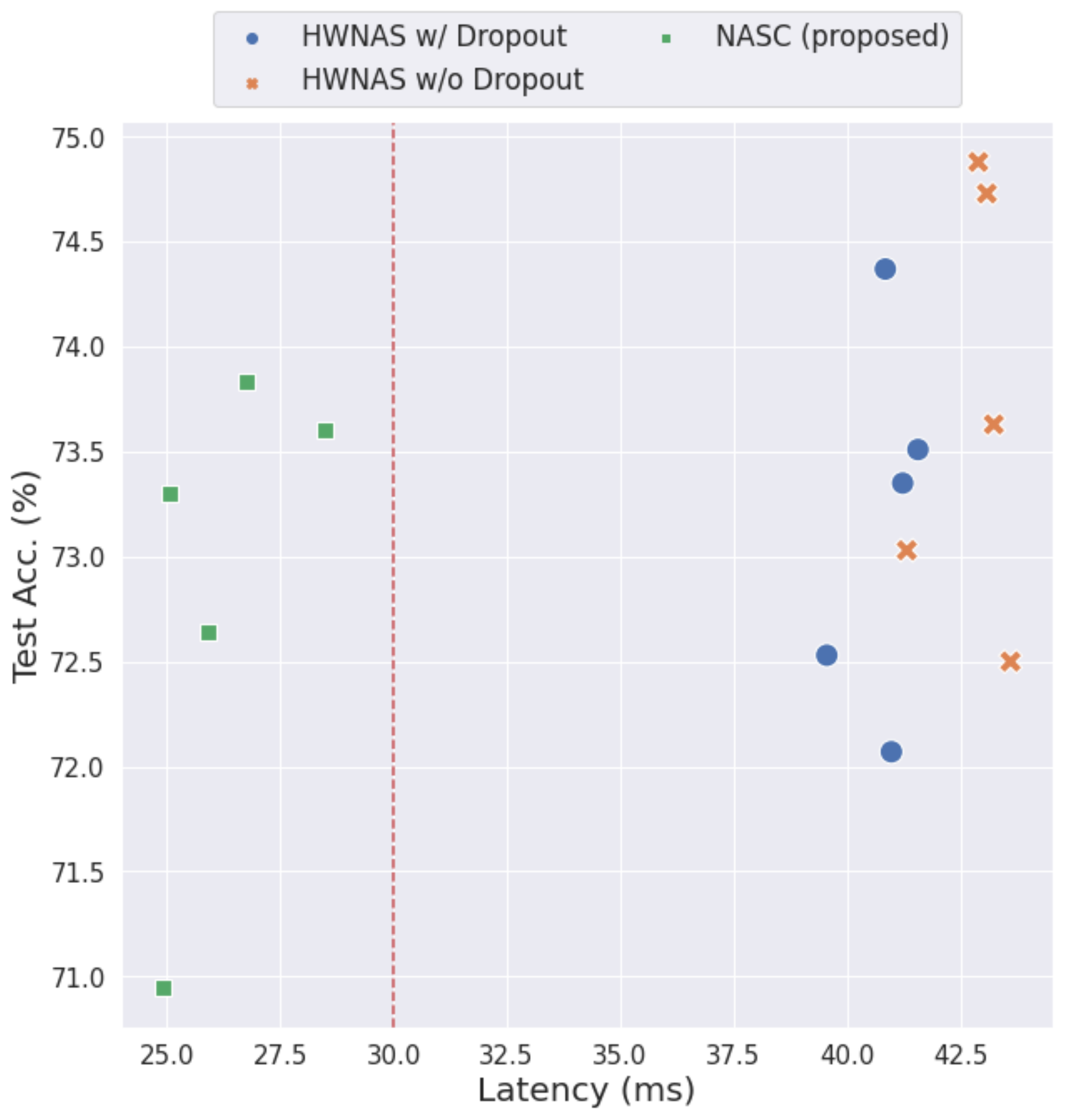}
        \caption{Accuracy vs. latency of each model searched in FBNet search space when $p=0.2$ and $T_\mathrm{th}=30$ ms.}
        \label{fig:FBNet}
\end{figure}

\vspace{3mm}
\noindent \textbf{Assumptions on communications:}\quad
An end device and an edge server were assumed to be connected via a lossy IoT network abstracted as a communication link in which the packets were randomly dropped with the probability $p$. Hence, a proportion $p$ of the elements of the intermediate representation (i.e., the output of the head network) transmitted by the device were randomly dropped. We also assumed a stable throughput $r$ for the communication link. Thus, the communication latency between the device and server is calculated as
\begin{align}
    T^\mathrm{comm} = \frac{D(a_\mathrm{h})}{r},
\end{align}
where $D(a_\mathrm{h})$ denotes the data size of the output of the head network $N_\mathrm{h}(\cdot | x_\mathrm{h}, a_\mathrm{h})$. The data size depends on the number of units in the output layer, the data type, and the compression method. In this evaluation, to calculate the communication latency simply, the data size is calculated as $q \times n_\mathrm{h}$ where $q$ and $n_\mathrm{h}$ represent the quantization bit rate and the number of output units in the head network, respectively. Assuming a data type of 32 bit float, $q$ was set to 32. The throughput of the communication link (including MAC and network layer overheads) $r$ was set to 8.0 Mbit/s, which is in the throughput range for wireless LANs based on the IEEE 802.11 standards. 



\vspace{3mm}
\noindent \textbf{Compared methods:}\quad 
We compared NASC with a hardware-aware NAS protocol modified for this split computing scenario. We refer to this protocol as HWNAS. The HWNAS employs a split point optimization method simplified from \cite{Li18} and model tuning leveraging dropout \cite{Itahara22} to reduce latency and improve robustness against packet loss. The HWNAS conducts conventional NAS, model split, and model re-training sequentially. Specifically, HWNAS first performs an architecture search assuming the entire model is on the end device; that is, it calculates the latency assuming $T = T_i^{\mathrm{comp}}(a_{\mathrm{h}}) + T_i^{\mathrm{comp}}(a_{\mathrm{t}})$ and assumes that packet loss does not occur in \eqref{eq:opt}. 
Then, the model is split into head and tail networks at a point that minimizes the same objective function \eqref{eq:opt} as NASC to minimize the latency and accuracy degradation by split computing. 
Finally, in the HWNAS w/ dropout, the head and tail networks are re-trained using the dropout technique, whereas the HWNAS w/o dropout simply re-trains the networks without dropout.

\begin{figure}
    \centering
    \includegraphics[width=0.7\linewidth]{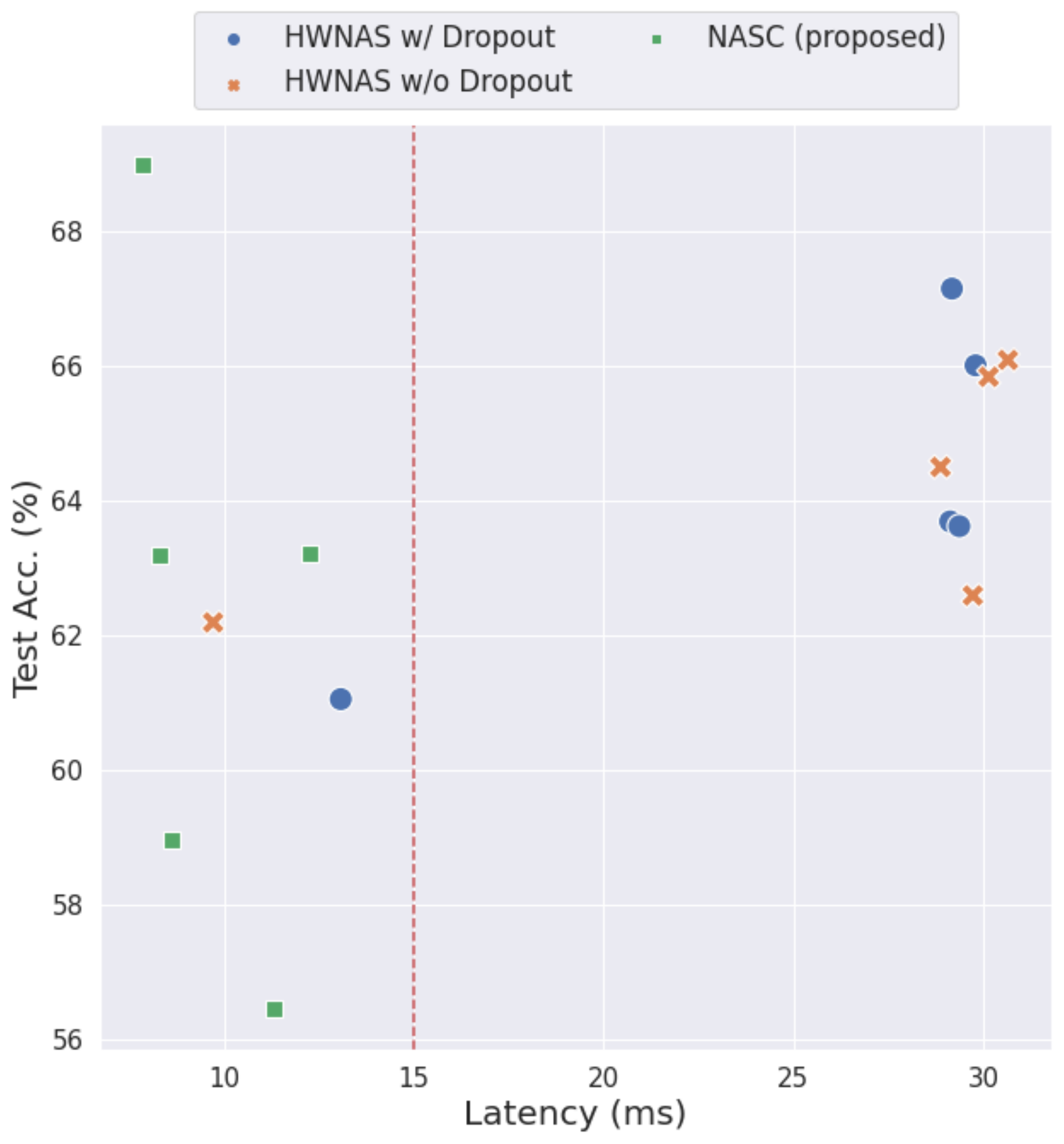}
    \caption{Accuracy vs. latency of each model searched in extended search space when $p=0.2$ and $T_\mathrm{th}=15$ ms.}
    \label{fig:original}
\end{figure}

\subsection{Results} 

Fig.~\ref{fig:FBNet} shows scatter plots of accuracy vs. latency for the 15 models obtained by the proposed method and the compared methods in the FBNet search space when $p=0.2$. Five models were obtained for each method. As shown in Fig.~\ref{fig:FBNet}, only the proposed method can obtain a model that satisfies the latency constraints. This is because that NASC searches a model that minimizes the total latency of computation and communication, while HWNAS does not consider communication latency when searching for model architectures. However, the model accuracy of NASC was slightly lower than HWNAS. Specifically, the medians of the model accuracy for NASC, HWNAS w/ Dropout, and HWNAS w/o Dropout were 73.30\%, 73.35\%, and 73.63\%, respectively. This is because of the trade-off between the model accuracy and latency.

Fig.~\ref{fig:original} shows scatter plots of the accuracy vs. latency when considering the extended search space when $p=0.2$. As in the FBNet search space, all the models obtained by the proposed method achieved lower latency than the threshold, whereas each baseline obtained only one model that achieves lower latency than the threshold. On the one hand, the median of the accuracy of NASC was slightly lower than baselines, which were 63.20\%, 63.69\%, and 64.50\% for NASC, HWNAS w/ Dropout, and HWNAS w/o Dropout, respectively.

These results demonstrate that NASC can significantly reduce latency while slightly decreasing model accuracy, thereby improving the accuracy-latency trade-off.

\section{Conclusion}
In this paper, we propose a NAS for split computing, called NASC. The NASC employs the adaptive stochastic natural gradient method to jointly explore the optimal model architecture and split point to achieve higher accuracy with low end-to-end latency, that is, to minimize the weighted sum of the model loss and total latency on communication and computation in the end device and edge server. The performance evaluation using HW-NAS-Bench demonstrates that the proposed NASC reduces the latency by approximately 40--60\% from the baseline with slight accuracy degradation.

\bibliographystyle{IEEEtran.bst}
\bibliography{NAS4SC.bib}

\end{document}